\documentclass{article}
\usepackage{float}

\usepackage[preprint]{neurips_2025}

\usepackage{booktabs}
\usepackage{multirow}
\usepackage{array}

\usepackage{xcolor}
\usepackage{colortbl}
\usepackage{float}
\usepackage{caption}
\usepackage{enumitem}
\usepackage[utf8]{inputenc} 
\usepackage[T1]{fontenc}    
\usepackage{hyperref}       
\usepackage{url}            
\usepackage{booktabs}       
\usepackage{amsfonts}       
\usepackage{nicefrac}       
\usepackage{microtype}      
\usepackage{xcolor}         
\usepackage{mathtools}
\usepackage{amsmath}
\usepackage{amssymb}
\usepackage{amsthm}

\usepackage{siunitx} 

\title{Adversarial Examples Are Not Bugs, They Are Superposition}

\author{%
  Liv Gorton\\
  Goodfire\\
  San Francisco, CA \\
  \texttt{liv@goodfire.ai}
  \And
  Owen Lewis\\
  Goodfire\\
  San Francisco, CA \\
  \texttt{owen@goodfire.ai} \\
}

\begin{document}

\maketitle

\begin{abstract}
Adversarial examples—inputs with imperceptible perturbations that fool neural networks—remain one of deep learning's most perplexing phenomena despite nearly a decade of research. While numerous defenses and explanations have been proposed, there is no consensus on the fundamental mechanism. One underexplored hypothesis is that \textit{superposition}, a concept from mechanistic interpretability, may be a major contributing factor, or even the primary cause. We present four lines of evidence in support of this hypothesis, greatly extending prior arguments by \cite{elhage2022superposition}: (1) superposition can theoretically explain a range of adversarial phenomena, (2) in toy models, intervening on superposition controls robustness, (3) in toy models, intervening on robustness (via adversarial training) controls superposition, and (4) in ResNet18, intervening on robustness (via adversarial training) controls superposition.
\end{abstract}


\section{Introduction}

Adversarial examples represent one of the most perplexing phenomena in deep learning: neural networks that achieve superhuman performance on many tasks can be fooled by perturbations so small they are imperceptible to humans. Despite nearly a decade of intensive research and many different hypotheses, there is no widely accepted explanation. In this paper, we explore an alternative hypothesis: superposition.

Superposition is a concept from the mechanistic interpretability literature. At a high level, superposition exploits the geometry of high-dimensional spaces to allow neural networks to represent more features than they have neurons. However, this strategy comes at a cost. Features in superposition necessarily interfere. On distribution, this interference is small, but in worst-case scenarios, it can be significant. One of the foundational papers on superposition hypothesized this interference could be linked to adversarial examples \citep{elhage2022superposition}, yet this hypothesis remains unexplored.

\begin{figure}[H]
    \centering
    \includegraphics[width=\textwidth]{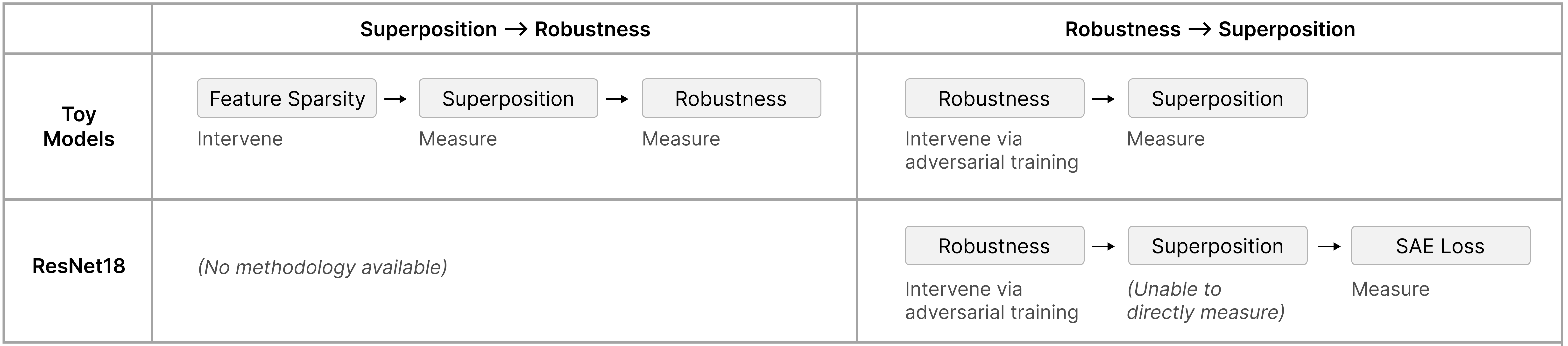}
    \caption{\textbf{Overview of experiments.} The three primary experiments test the relationship between superposition and robustness in different ways. }
    \label{fig:argument_summary}
\end{figure}

Our primary contribution is three experiments testing the relationship between superposition and robustness, in both toy models an ResNet18. These experiments are summarized in Figure \ref{fig:argument_summary}. For toy models, we demonstrate both that superposition can control robustness, and that robustness can control superposition. For ResNet18, we show only that robustness can control superposition. (Unfortunately, without a method for controlling superposition in real models, we are unable to demonstrate the other direction in real models.)

Combined, these results strongly imply that superposition is at least one causal factor in the existence of adversarial examples. They don't necessarily suggest that it's the only factor, as we can't intervene on superposition in real models to isolate this. 

At the same time, we take seriously the possibility that it might be the primary explanation. Although it isn't the primary focus of this paper, it seems to us that superposition is sufficient to theoretically explain all the adversarial phenomena we're aware of. This is summarized in Table \ref{table:phenomena}. 



\begin{table}[H]
    \centering \small
    \renewcommand{\arraystretch}{1.3}
     \caption{Six adversarial example phenomena and potential explanations.}
     \begin{tabular}{{|p{6.5cm}|p{6.5cm}|}}
    \hline
        \textbf{Phenomenon} & \textbf{Superposition Explanation} 
        \\ \hline \hline
         \textbf{Existence:} Adversarial examples exist across essentially all neural networks \citep{szegedy2014intriguingpropertiesneuralnetworks, goodfellow2015explainingharnessingadversarialexamples}&  Features can be attacked by perturbing all the features in superposition with them. An attacker can do this iteratively at each layer.  \\ \hline
          \textbf{Noise-like structure:} Adversarial perturbations appear as unstructured high-frequency noise rather than semantic patterns \citep{goodfellow2015explainingharnessingadversarialexamples, sharma2019effectivenesslowfrequencyperturbations}& Adversarial attacks work by attacking many features, which are totally unrelated except for the fact that they're in superposition with the actual targets. \\\hline
         \textbf{Attack Transferability:} Adversarial examples transfer between independently trained models \citep{goodfellow2015explainingharnessingadversarialexamples, liu2017delvingtransferableadversarialexamples}&  If the same features are in superposition with each other, attacks based on superposition will transfer.  Features which are anti-correlated are preferentially put in superposition with each other \citep{elhage2022superposition} and therefore attacks should transfer. \\\hline
         \textbf{Training difficulty:} Adversarial training is fundamentally difficult, requiring significant computational resources and degrading natural accuracy \citep{madry2019deeplearningmodelsresistant}& Superposition increases the capacity of models. If improving model robustness requires reducing superposition, that fundamentally reduces model capacity.\\\hline
         \textbf{Interpretability:} Adversarially trained models become markedly more interpretable with neurons that correspond to human-understandable concepts \citep{engstrom2019adversarialrobustnesspriorlearned} & In the absence of superposition, neurons can be monosemantic, and also less noisy.\\\hline
         \textbf{Training on Attacks Transfers Clean Performance:} Training on \emph{mislabeled} data with adversarial attack towards the erroneous label induces correct behavior on clean data \citep{ilyas2019adversarialexamplesbugsfeatures}& Training on adversarial attacks transfers to clean data because adversarial attacks encode interfering combinations of genuinely useful circuits.\\\hline

    \end{tabular}
    
    \label{table:phenomena}
\end{table}

\section{Background}

The mechanistic interpretability literature often assumes that model representations are linear. That is, the hidden activations $h$ of some layer can be understood as $$h=\sum_{i<k} a_i \vec{f_i} + \vec{b} $$ where $k$ is the total number of features, $a_i$ is the activation of a feature $i$, and $f_i$ is a direction in activation space representing that feature. Roughly, activation represents the intensity or strength of a feature in response to a particular input.\footnote{Typically, features are imagined to be one-dimensional, but this can be generalized to allow more dimensions.} 


One might expect that if a neural network representation has $n$ dimensions, it can only represent $k\leq n$ linear features. However, results from an area of mathematics called compressed sensing suggest that neural networks could represent many more features ($k>>n$), so long as features are sparse (that is, zero on most examples). This is called the superposition hypothesis.

Superposition necessarily entails \textit{interference}. When $k > n$ features are represented in an $n$-dimensional space, the feature vectors $\{\vec{f}_i\}_{i=1}^k$ cannot all be mutually orthogonal. This non-orthogonality means that activating feature $i$ with coefficient $a_i$ produces (apparent) spurious activations in feature $j$ proportional to $a_i \langle \vec{f}_i, \vec{f}_j \rangle$. Models can partially compensate for this interference by learning negative biases $b_j < 0$ that suppress small spurious activations below a threshold. However, this compensation mechanism assumes the total interference $\sum_{i \neq j} a_i \langle \vec{f}_i, \vec{f}_j \rangle$ remains bounded. In worst-case scenarios, an adversary can coordinate activations to make this sum arbitrarily large, overwhelming the bias term. (This aligns with compressed sensing theory, which only guarantees reconstruction with high probability under random, not adversarial, conditions.)

\cite{elhage2022superposition} demonstrated that this interference mechanism enables adversarial attacks in toy models. Specifically, consider a target feature $\vec{f}_{\text{target}}$ in superposition with features $\{\vec{f}_1, \ldots, \vec{f}_m\}$ where $\langle \vec{f}_{\text{target}}, \vec{f}_i \rangle = \epsilon_i \neq 0$. An adversary can exploit this by adding input perturbations that activate each interfering feature by a small amount $\delta_i$. While each individual contribution $\delta_i \epsilon_i$ to the target feature's activation is negligible, the cumulative effect $\sum_{i=1}^m \delta_i \epsilon_i$ can be made arbitrarily large by choosing appropriate $\delta_i$ values (subject to the perturbation budget). This is precisely the interference that models attempt to suppress through learned biases under normal operating conditions.

This vulnerability compounds across layers. At each layer, the adversary can exploit superposition to create unwanted feature activations, which then propagate to the next layer as inputs. These corrupted activations at the next layer can then be constructed to do the same kind of attack, allowing errors to accumulate through the network.

\section{Causal Evidence from Toy Models of Superposition}

To test whether superposition causally contributes to adversarial vulnerability, we extend the toy models of \cite{elhage2022superposition}, the standard theoretical model of superposition. In the toy model setup, it is possible to exactly measure superposition, which is not possible in real models because it requires knowledge of the ground truth features learned by the model. It also allows us to control superposition by manipulating feature sparsity. This will allow us to show both that superposition controls robustness and that robustness controls superposition in the toy models setup.

\subsection{Setup}

\subsubsection{Toy Models}\label{section:toy_models}

We consider a simplified\footnote{We consider only uniform feature importance, causing the loss to simplify into mean squared error.} version of the basic setup of \cite{elhage2022superposition}. Our data consists of $n=100$ features. They are linearly projected into a  $m=20$ hidden units, $h=Wx$, and then reconstructed by a ReLU layer, $x' = \text{ReLU}(W^T x + b)$. The loss is mean squared error.  

The behavior of this toy model varies based on the feature sparsity, $S$. This is the probability that the input features are zero. When features are sparse, this setup exhibits superposition, representing more features than there are hidden dimensions.  The amount of superposition increases with sparsity.

\subsubsection{Measuring Superposition}
One reason for our interest in the toy model setting is that superposition can be exactly measured. One way to do this is by looking at the features per dimension \citep{elhage2022superposition}, i.e., how many features the model is attempting to represent per feature dimension: 
\begin{equation} \label{eq:features_per_dim}
\frac{||W||_F^2}{n}
\end{equation}

This works because features are roughly represented with unit norm when learned. When the features per dimension $>1$, the model must be using superposition, as it represents more features than it has dimensions. 




\subsubsection{Measuring Robustness}

We also need to know how vulnerable our models are to adversarial examples. To measure adversarial vulnerability, we generate $L_2$-bounded adversarial examples. For each input $x$, we find the worst-case perturbation within an $\epsilon$-ball that maximizes reconstruction error: 
\begin{equation}
x_{adv} = x + \epsilon \cdot \arg\max_{\|\delta\|_2 \leq 1} \mathcal{L}(x + \epsilon \delta)
\end{equation}
We set $\epsilon$ to 10\% of the average input norm.

We reproduce the approach of \citet{elhage2022superposition}, who exploit the toy model setup to analytically construct attacks that optimally attack each specific output feature, and then take the worst such attack. They take this approach to avoid gradient masking issues from ReLU. However, while this would be an optimal attack in terms of $L_\infty$ in the output space, it has the potential to be quite suboptimal for affecting the output as measured by $L_2$/MSE. For this reason, we primarily consider a more traditional adversarial attack. We add a small amount of noise to avoid gradient issues, and then do a one-step gradient L2 attack. All results in the main paper are based on this attack.

To compare the vulnerability of models, we consider how many times more vulnerable it is than a model without superposition (i.e., our model with the highest input feature density, with every feature present in all training inputs).

\subsubsection{Adversarial Training Protocol}\label{section:tms_adv_training}

Since we want to test whether causality flows from adversarial robustness to superposition, we also need to be able to produce adversarially robust versions of our toy models. To do this, we train new toy models over the same range of feature densities, but using a mixture of clean and adversarial training examples:
\begin{equation}
\mathcal{L}_{adv} = \alpha \cdot \mathcal{L}(x) + (1-\alpha) \cdot \mathcal{L}(x_{adv})
\end{equation}
where $\alpha = 0.5$ balances clean and \textit{robust} accuracy. We use $L_2$ attack with $\epsilon = 0.1 \|x\|_2$. We can generate these attacks on-the-fly using either approach from the previous section, but unless otherwise specified, we use the more standard gradient attack rather than the Elhage method. We train a model with the same configuration as the model used in Section \ref{section:toy_models} for 150,000 steps with a learning rate $10^{-3}$. (This follows a common practice in adversarial training where models are trained for extended periods compared to standard training due to the unique optimization dynamics; see e.g., \citet{rice2020overfittingadversariallyrobustdeep} for discussion of adversarial training dynamics.)

\subsection{Intervening on Superposition Controls Adversarial Vulnerability}

We use feature sparsity to manipulate the level of superposition, and observe resulting changes in adversarial robustness. In particular, we vary the feature density $(1 - \text{sparsity})$ exponentially from 1.0 to 0.1, training 30 models simultaneously with different sparsity levels, and observe the resulting adversarial robustness. This is the general setup of \citep{elhage2022superposition}, but we focus on more powerful noise-plus-gradient adversarial attacks. (A reproduction of the original Elhage experiment can be found in the appendix, see figure \ref{fig:tms_replication}.)

Our first goal is to confirm that intervening on feature sparsity has the expected effect on superposition, in order to validate it as a way to manipulate superposition in our larger experiment. Panel A of figure \ref{fig:tms_adversarial_training} shows the expected results, including a temporary plateau corresponding to antipodal superposition.

Having validated our instrumental variable, we now proceed to the core result. Panel B of figure \ref{fig:tms_adversarial_training} shows that adversarial vulnerability increases with both feature sparsity and superposition (quantified as features per dimension). There is one striking dip corresponding to antipodal superposition.

The mechanism is intuitive: when features are in superposition, they share directions in activation space. An adversary can exploit this by perturbing all interfering features simultaneously. Since features in superposition are not orthogonal, small perturbations to many features accumulate into large changes in the target feature's reconstruction.

It is worth noting that there is some subtlety to comparing adversarial robustness across different feature densities, since the distribution we are evaluating on changes. However, this should, if anything, bias in the opposite direction of the trend we're observing. Having fewer features active should tend to make models more robust, since fewer ReLUs would be open, allowing gradients through. Thus, we believe this concern would cause us to \textit{underestimate} the relationship between superposition and adversarial vulnerability. However, we do get some cross-validation from the robust models in the next section, since these shift superposition independently of the data distribution, and we still see the same trend.

\subsection{Intervening on Adversarial Robustness Controls Superposition}

To establish bidirectional causality, we next ask: does improving adversarial robustness reduce superposition? We perform adversarial training on our toy models and measure the resulting changes in superposition.

\begin{figure}[H]
\centering
\includegraphics[width=\textwidth]{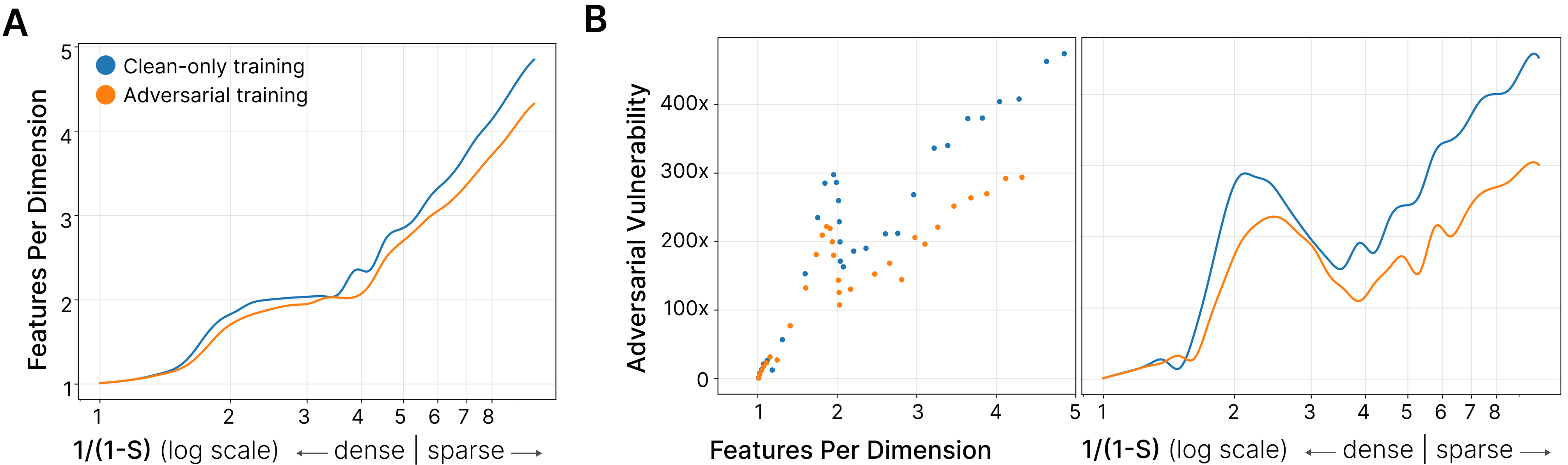}
\caption{\textbf{Adversarial training reduces superposition.} Comparison of models before (blue) and after (orange) adversarial training. \textbf{A:} Features per dimension decreases for a given sparsity level. \textbf{B:} (Left) Models become more vulnerable to adversarial examples as superposition increases. (Right) Models become more vulnerable to adversarial examples as feature sparsity increases (with a drop for antipodal superposition). }
\label{fig:tms_adversarial_training}
\end{figure}

Figure \ref{fig:tms_adversarial_training} demonstrates that adversarial training reduces superposition. Models that underwent adversarial training decreased their adversarial vulnerability and decreased features per dimension for some original input sparsity. However, we note two surprising phenomena. Firstly, as discussed earlier, we note a drop in vulnerability to adversarial examples when models switch to antipodal superposition. Secondly, we note that robust models are often more robust than expected for their superposition level. Our interpretation is that the overall level of superposition doesn't tell the full story; we conjecture that some superposition structures (that is, the matrix of interference between features) are more or less vulnerable to superposition. See Discussion (section \ref{section:discussion}).

In contrast to the previous section, where we reproduced and extended the results of \cite{elhage2022superposition}, to the best of our knowledge, these results are the first to demonstrate causality from robustness to superposition.

\subsubsection{Theoretical Intuition}

While not a formal derivation, we find it useful to conceptualize the difference between standard and adversarial training through the lens of interference minimization.\footnote{This is a conceptual framework for building intuition rather than a formal theoretical result. The actual optimization dynamics are considerably more complex.}

Given dataset $\mathcal{D}$, neural network parameters $\theta$, and a measure of interference $I$, we might conceptualize neural network training as:
\begin{equation*}
    \min_\theta \mathbb{E}_{(x, y) \sim \mathcal{D}}[I(x, y; \theta)]
\end{equation*}

That is, the goal is to minimize the \textit{average} expected interference. Whereas during adversarial training, it might be better instead to conceptualize the objective with respect to interference as:
\begin{equation*}
\min_\theta \max_{\mathcal{D} \in \mathcal{D}_{\text{OOD}}} \mathbb{E}_{(x, y) \sim \mathcal{D}}[I(x, y; \theta)]
\end{equation*}

That is, the goal is to minimize the \textit{maximum} expected interference over out-of-distribution data.

This conceptualization suggests that adversarial training forces the model to consider worst-case interference patterns rather than average-case, potentially explaining why it reduces superposition in our experiments.

\subsubsection{Adversarial Examples Exploit Feature Interference}\label{section:feature_interference_graphs}

We constructed superposition geometry graphs similarly to \citet{elhage2022superposition}, where each feature has a node, and edge $(i, j)$ represents $(W_i \cdot W_j)^2$.

These graphs can then be used to understand how this geometry is being exploited in adversarial attacks, and subsequently why a model is adversarially robust.


\begin{figure}[h]
\centering
\includegraphics[width=\textwidth]{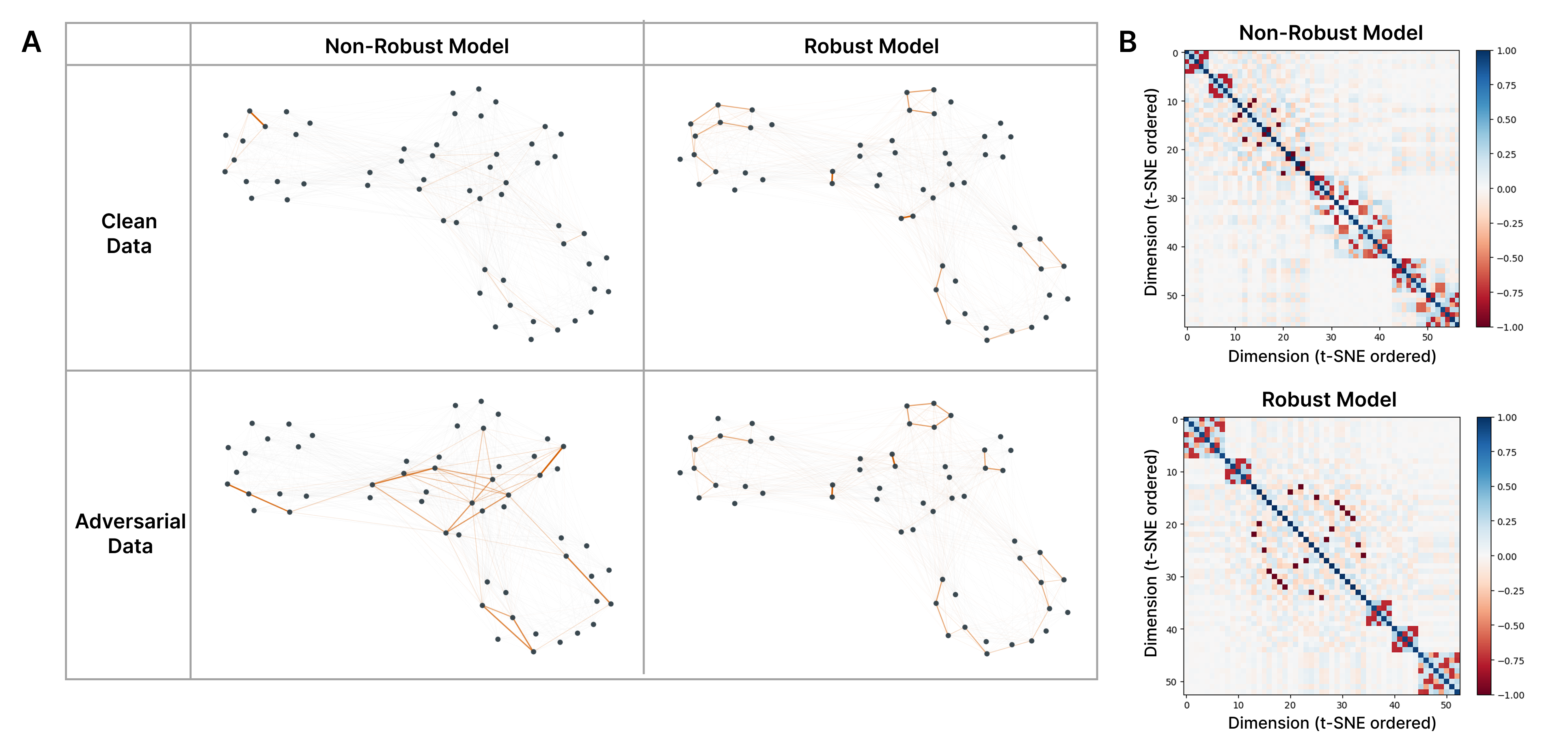}
\caption{\textbf{Adversarial attacks activate interfering features in superposition.} (A) We consider two models, one robust and one non-robust, as well as clean and adversarial data. We visualize the superstructure of each toy model as a graph. Edge thickness is dependent on $(W_i \cdot W_j)^2$. We then highlight the superposition affecting that input in orange. (B) We plot heatmaps of the interference ($W^\top W$) for the robust and non-robust models used in (A). Non-robust models have a mean off-diagonal interference $2\times$ that of robust models.}
\label{fig:tms_graph_feature_interference}
\end{figure}

Figure \ref{fig:tms_graph_feature_interference} illustrates how adversarial attacks exploit feature interference patterns. In non-robust models (left column), clean inputs activate relatively few features with minimal interference between them, as shown by the sparse orange highlighting in the superposition graph. Adversarial inputs, however, activate many interfering features simultaneously, precisely the pattern expected if attacks exploit superposition geometry. In contrast, robust models (right column) show similar sparse activation patterns for both clean and adversarial inputs, suggesting that adversarial training has reorganized the feature geometry to prevent interference-based attacks. The heatmaps in panel (B) confirm this: non-robust models exhibit mean off-diagonal interference approximately 2× that of robust models, indicating denser superposition structure.
 
\subsection{Superposition Geometry}\label{section:superposition_geometry}

\begin{figure}[h]
\centering
\includegraphics[width=\textwidth]{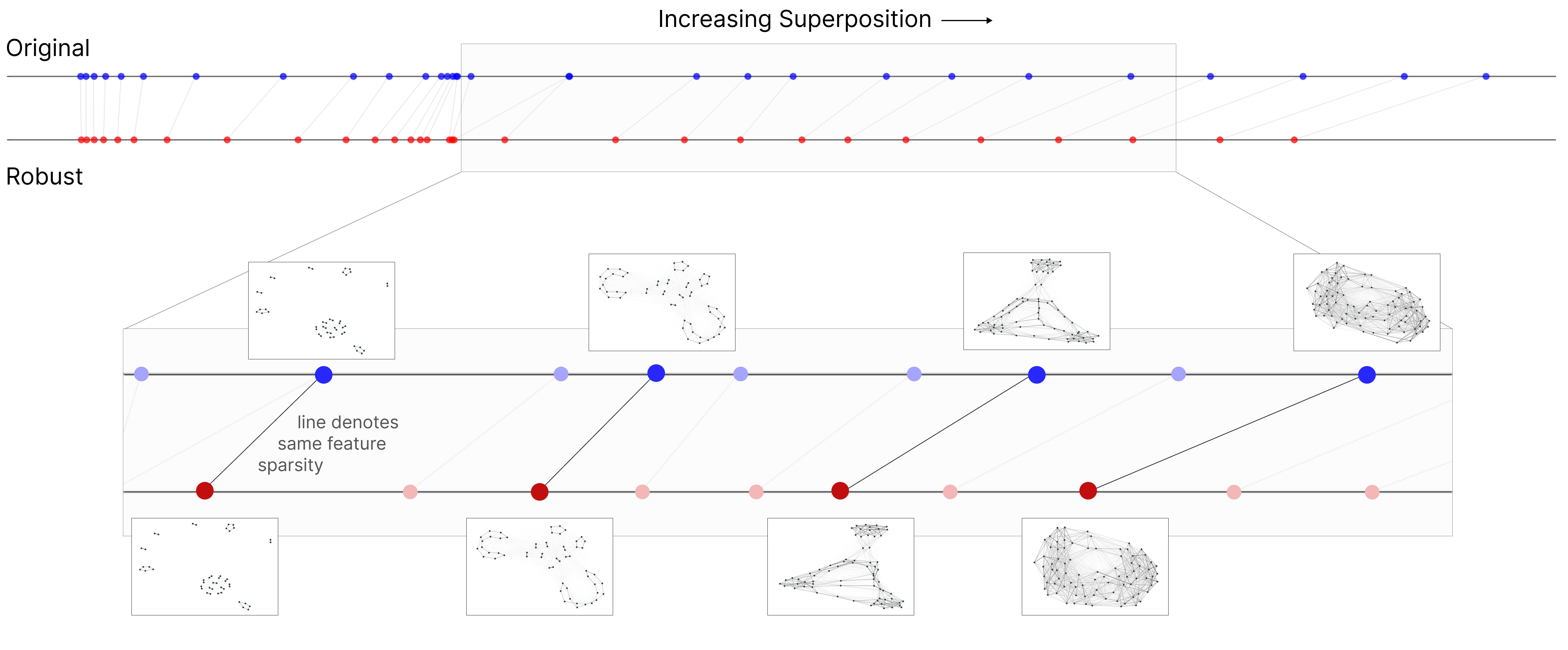}
\caption{\textbf{Adversarial training reduces superposition while preserving geometric structure.} We plot all our robust and non-robust models as "points on a superposition number line". A line connects models trained on the same level of sparsity. We can see that robust models have lower superposition. For selected models, we visualize the superposition structure as a graph.}
\label{fig:tms_graphs_fpd}
\end{figure}

We can also use the graph visualization technique to compare a larger set of models. In figure \ref{fig:tms_graphs_fpd}, we look at pairs of non-robust and robust models trained at the same sparsity level. The robust models have less superposition (corresponding to a further left position) but strikingly similar superposition geometries.

\section{Evidence From Real Models}

We now turn our attention to real models. Unfortunately, since we have no way to intervene on superposition in real models, we can't test the causal effect of superposition on robustness. However, we can still adversarially train models to control robustness and observe the effect on superposition via the proxy of sparse autoencoder loss (discussed further in Section \ref{section:sae_reconstruction})

\subsection{Methods}

\subsubsection{Adversarially Robust Models}

To study adversarial robustness in real models, we used robust ResNet18s trained on ImageNet \citep{imagenet15russakovsky} from \citet{salman2020adversariallyrobustimagenetmodels}.\footnote{\url{https://huggingface.co/madrylab/robust-imagenet-models}} These robust models are trained against different attack sizes, varying their robustness.

\subsubsection{Sparse Autoencoders}

We train sparse autoencoders (SAEs) on the outputs of ResNet18's four residual stages (conv2\_x through conv5\_x), which produce 256-, 512-, 1024-, and 2048-dimensional feature maps at progressively lower spatial resolutions. We trained both L1 ReLU SAEs \citep{anthropic2024aprilupdate} and TopK SAEs \citep{gao2024scalingevaluatingsparseautoencoders} on standardized activations to mitigate the effect on training of activation statistics. Additional training details can be found in appendix \ref{appendix:sae_training}.

\subsection{Robust Models Achieve Better SAE Reconstruction}\label{section:sae_reconstruction}

There is no direct way to measure the amount of superposition in real models, and so instead we must consider proxies of superposition.

SAEs are designed to model superposition and will naturally have a higher loss when there is more superposition. There are several reasons for this: (1) if a model of a fixed size has more superposition, it has more total features that an SAE has to model, (2) with more total features, there will also be more active features on any example, (3) in denser superposition, the SAE will be forced to either sometimes model a strongly activating feature as activating other features, or sometimes not represent small activations.

\begin{figure}
\centering
\includegraphics[width=0.5\textwidth]{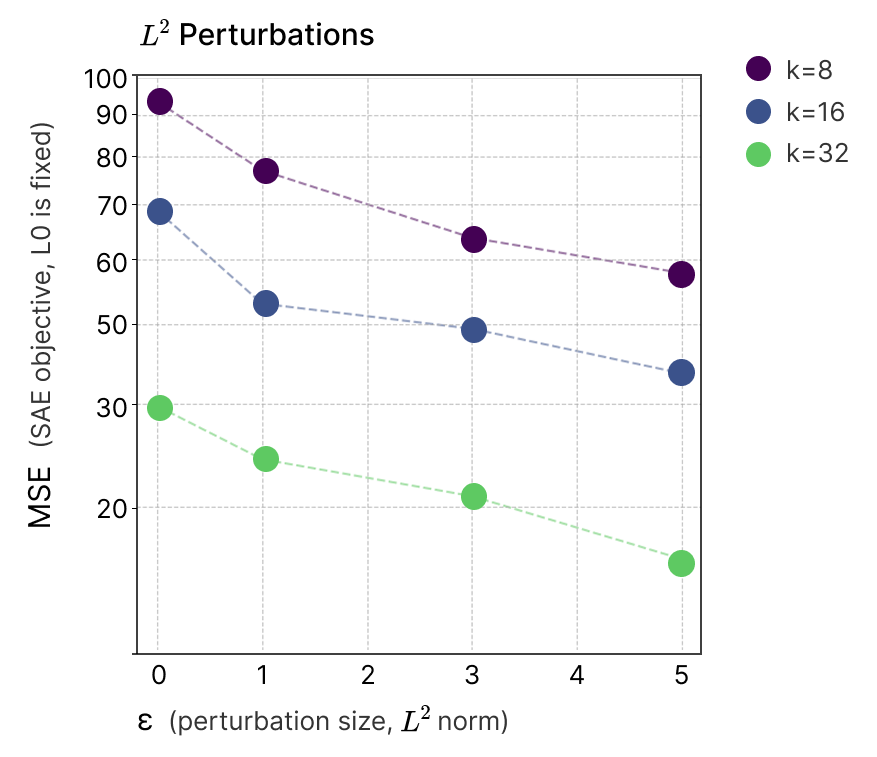}
\caption{\textbf{Robust models achieve better reconstruction at a given sparsity level.} TopK SAEs with different sparsity levels ($k=\{8, 16, 32\}$) were trained on ResNet18 models with varying L2 robustness ($\epsilon \in \{0, 1, 3, 5\}$). Lower MSE at fixed sparsity likely indicates less interference and therefore less superposition.}
\label{fig:resnet18_topk_mse}
\end{figure}

Figure \ref{fig:resnet18_topk_mse} shows that for a given sparsity level, more robust models consistently achieve better reconstruction loss. Does this imply robustness effects superposition? The only way we see to avoid this is if some other change to the model could lower SAE loss independent of superposition, and we don't have any hypotheses for what that could be.\footnote{From a Popperian perspective, the hypothesis that robustness influences superposition should gain credit for predicting a surprising phenomenon, even if some alternative explanation can retrospectively be proposed.}

\subsection{Adversarial Examples Increase L0}

Our sparse autoencoders provide the opportunity for an additional experiment. If adversarial attacks do exploit interference, we'd expect them to activate more features. Each feature can both be attacked via interference and used to attack later features.

\begin{figure}[H]
\centering
\includegraphics[width=\textwidth]{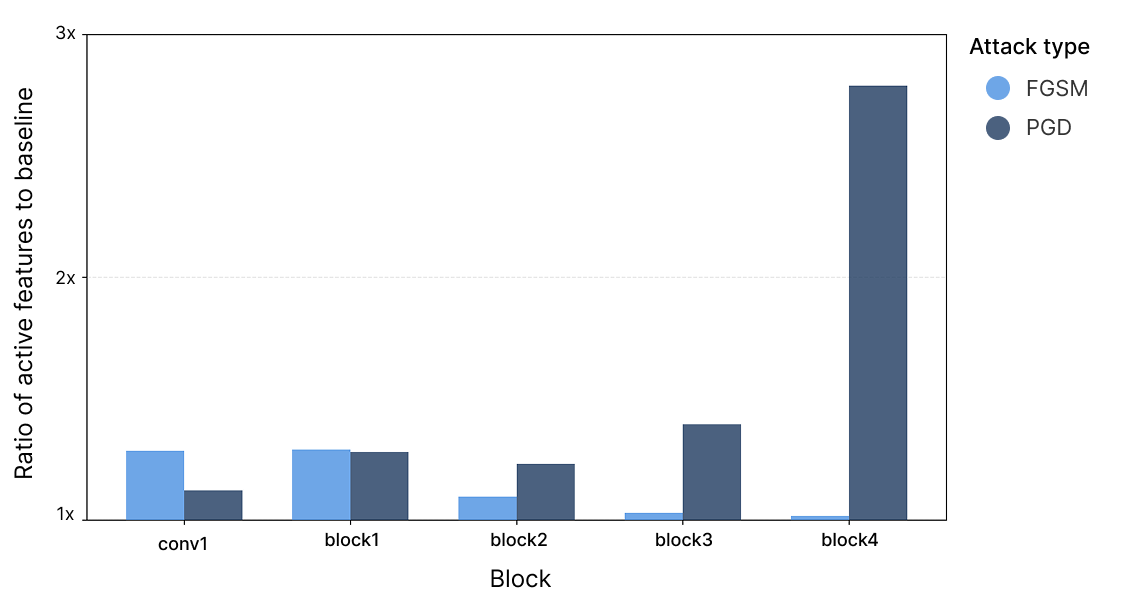}
\caption{\textbf{Adversarial examples activate more features than clean inputs.} Ratio of L0 (active features) for FGSM \citep{goodfellow2015explainingharnessingadversarialexamples} and PGD \citep{madry2019deeplearningmodelsresistant} attacks versus clean data across ResNet18 layers. PGD shows a dramatic increase at layer 4 (2.8$\times$), suggesting adversarial attacks can increasingly exploit feature interference deeper in the network. See Table \ref{tab:l0_statistics} for detailed statistics.}
\label{fig:resnet18_L0}
\end{figure}

In figure \ref{fig:resnet18_L0}, we observe that adversarial examples consistently activate more features than clean inputs across all layers. For FGSM attacks, we see modest increases of 1.3$\times$ at conv1, maintaining similar levels through layers 1-3, before dropping to near baseline (1.02$\times$) at layer 4. PGD attacks show a different pattern: starting with 1.1-1.3$\times$ increases in early layers (conv1 and layer1), maintaining moderate increases through layers 2-3 (1.2-1.4$\times$), then dramatically spiking to 2.8$\times$ at layer 4. This striking divergence between attack types at the final layer suggests that iterative attacks (PGD) can more effectively exploit accumulated interference in deeper representations. This aligns with the well-established finding that PGD, as a stronger multi-step optimization-based attack, typically achieves higher success rates than single-step methods like FGSM \citep{madry2019deeplearningmodelsresistant}.


\section{Discussion}\label{section:discussion}

We have argued that adversarial examples are caused, at least in part, by superposition. Beyond the theoretical arguments, three lines of empirical evidence support this hypothesis: (1) in toy models, superposition controls robustness, (2) in toy models, robustness controls superposition, and (3) in real models, robustness controls superposition.

While these arguments appear compelling, several limitations warrant consideration. First, our analysis relies substantially on proxy variables to control and measure effects, particularly in real models. These proxies may fail to capture the full complexity of the phenomena. Second, our experimental results could be consistent with adversarial examples having multiple causal factors beyond superposition, especially in real models. Without methods to directly manipulate superposition in real models and observe resulting changes in robustness, we cannot quantify the relative magnitude of superposition's contribution, only establish the potentiality of a causal relationship. Despite these limitations, the evidence strongly suggests that superposition constitutes a major factor in adversarial robustness. Further confidence in this hypothesis will require developing more sophisticated tools for measuring and manipulating superposition in real models.

Several unexpected findings merit further investigation: (1) the temporary improvement in robustness observed near antipodal superposition configurations, and (2) the observation that models with equivalent overall superposition levels but different superposition structures exhibit varying robustness to L2 adversarial attacks. These phenomena warrant deeper theoretical and empirical examination.

If superposition represents a primary cause of adversarial examples, this implies a fundamental and unavoidable trade-off. Superposition enables models to effectively simulate substantially larger sparse models; achieving robustness would necessitate sacrificing this computational advantage. Conversely, this relationship would indicate a profound alignment between the objectives of interpretability and robustness research.

\section{Related Work}

\textbf{Adversarial Examples.} Since their discovery \citep{szegedy2014intriguingpropertiesneuralnetworks, goodfellow2015explainingharnessingadversarialexamples}, numerous attacks emerged \citep{DBLP:journals/corr/Moosavi-Dezfooli15, DBLP:journals/corr/CarliniW16a, madry2019deeplearningmodelsresistant, DBLP:journals/corr/abs-2003-01690}, extending to physical \citep{DBLP:journals/corr/KurakinGB16} and universal perturbations \citep{moosavidezfooli2017universaladversarialperturbations}. 

\textbf{Theoretical Explanations.} Beyond the linear hypothesis \citep{goodfellow2015explainingharnessingadversarialexamples}, explanations include geometric perspectives \citep{gilmer2018adversarialspheres, khoury2019on, shafahi2020adversarialexamplesinevitable, shamir2022dimpledmanifoldmodeladversarial}, concentration of measure \citep{mahloujifar2018curseconcentrationrobustlearning, mahloujifar2019empiricallymeasuringconcentrationfundamental}, high-dimensional inevitability \citep{tanner2024highdimensionalstatisticalmodel}, and manifold analyses \citep{xiao2022understandingadversarialrobustnessonmanifold}. The "robust features" hypothesis \citep{ilyas2019adversarialexamplesbugsfeatures} suggests models exploit non-robust but predictive patterns.

\textbf{Defenses.} Adversarial training remains dominant \citep{madry2019deeplearningmodelsresistant, zhang2019theoreticallyprincipledtradeoffrobustness, shafahi2019adversarialtrainingfree}, while certified approaches use verification \citep{zhang2018efficientneuralnetworkrobustness, gowal2019effectivenessintervalboundpropagation, wang2021betacrownefficientboundpropagation} or randomized smoothing \citep{cohen2019certifiedadversarialrobustnessrandomized, lecuyer2019certifiedrobustnessadversarialexamples}.

\textbf{Robustness-Accuracy Tradeoff.} Fundamental tension exists between standard and robust accuracy \citep{tsipras2019robustnessoddsaccuracy, zhang2019theoreticallyprincipledtradeoffrobustness, javanmard2020precisetradeoffsadversarialtraining, rice2020overfittingadversariallyrobustdeep, schmidt2018adversariallyrobustgeneralizationrequires}, with mitigations via unlabeled data \citep{carmon2022unlabeleddataimprovesadversarial, raghunathan2020understandingmitigatingtradeoffrobustness}.

\textbf{Interpretability.} Robust models exhibit aligned gradients and interpretable features \citep{engstrom2019adversarialrobustnesspriorlearned, tsipras2019robustnessoddsaccuracy, ganz2023perceptuallyalignedgradientsimply, srinivas2024modelsperceptuallyalignedgradientsexplanation}; disentangled representations improve robustness \citep{Yang_Guo_Wang_Xu_2021, guesmi2024exploringinterplayinterpretabilityrobustness}.

\textbf{Transferability and Compression.} Examples transfer due to shared representations \citep{demontis2019adversarialtransfer, wu2018understandingenhancingtransferabilityadversarial}; compression-robustness connections reveal capacity constraints \citep{ye2021adversarialrobustnessvsmodel, gui2019modelcompressionadversarialrobustness, xie2019informationtheoreticexplanationadversarialfragility, yi2020derivationinformationtheoreticallyoptimaladversarial}.

\textbf{Superposition and Mechanistic Interpretability.} Superposition allows exponentially many features in high-dimensional spaces \citep{elhage2022superposition}. SAEs decompose superposed features \citep{cunningham2023sparseautoencodershighlyinterpretable, bricken2023monosemanticity, templeton2024scaling, gao2024scalingevaluatingsparseautoencoders}, though computational bounds exist \citep{adler2025complexityneuralcomputationsuperposition}.

\section*{Acknowledgments} 

We would like to thank Chris Olah for helpful discussions that contributed to the development of this work. We are also grateful to Michael Byun, Tom McGrath, and Michael Pearce for their feedback on drafts of this manuscript.

\bibliographystyle{plainnat}  
\bibliography{references}      

\appendix

\section{Toy Models of Superposition Replication}\label{appendix:tms_replication}

\begin{figure}[H]
    \centering
    \includegraphics[width=\textwidth]{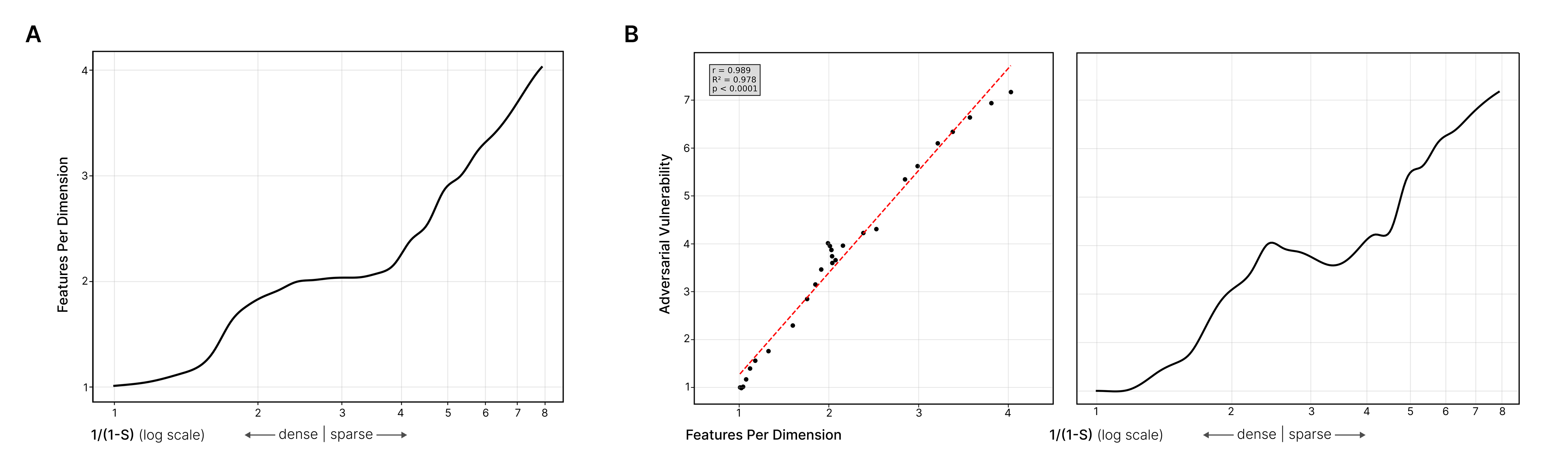}
    \caption{\textbf{Sparsity controls superposition, which drives adversarial vulnerability.} (A) Features per dimension increases with sparsity level 1/(1-S), with phase transitions at $\sim 1.7$ and $\sim 4$ corresponding to the onset of superposition and beyond-antipodal arrangements \cite{elhage2022superposition}. (B) Left: Adversarial vulnerability increases with feature sparsity. Right: Direct correlation between superposition (features per dimension) and adversarial vulnerability ($r \approx 0.99, p < 0.0001$). Each point represents a model trained at different sparsity. Results shown for Elhage-style attacks; see Figure \ref{fig:tms_adversarial_training} for gradient-based attacks.}
    \label{fig:tms_replication}
\end{figure}

\section{Sparse Autoencoder Training Details}\label{appendix:sae_training} 

All SAEs were trained with a batch size of 4096, a learning rate of $5\times10^{-4}$, and an expansion factor of $8\times$. Activations from models trained with different epsilons had slightly different distributions. Thus, for SAE training, activations were standardized using the mean and standard deviation for that specific model computed over a subset of the training data. 

When training TopK SAEs, top-$k_{\mathrm{aux}}$ was 512 and the auxiliary loss weight was 1.  

\pagebreak

\section{Supplementary L0 Statistics}

\begin{table}[h]
\centering
\caption{L0 activation values (mean ± SEM) for clean and adversarial images across network layers. Statistics computed from n=100,000 images per condition.}
\begin{tabular}{llll}
\toprule
Layer & Clean L0 & FGSM L0 & PGD L0 \\
\midrule
conv1 & 35.958 ± 0.0256 & 46.201 ± 0.0201 & 40.353 ± 0.0230 \\
layer1 & 32.876 ± 0.0195 & 42.426 ± 0.0102 & 42.092 ± 0.0081 \\
layer2 & 61.630 ± 0.0266 & 67.601 ± 0.0151 & 75.876 ± 0.0097 \\
layer3 & 72.798 ± 0.0341 & 74.987 ± 0.0282 & 101.469 ± 0.0227 \\
layer4 & 126.016 ± 0.0680 & 128.128 ± 0.0643 & 351.368 ± 0.1148 \\
\bottomrule
\end{tabular}
\label{tab:l0_statistics}
\end{table}

\section*{Revision History}

\paragraph{15th September, 2025} There was a bug in the plotting code for Figure \ref{fig:resnet18_L0} increasing the difference between clean and adversarial images which has now been fixed.

\end{document}